\documentclass{article}

\usepackage{PRIMEarxiv}

\usepackage[utf8]{inputenc} % allow utf-8 input
\usepackage[T1]{fontenc}    % use 8-bit T1 fonts
\usepackage{hyperref}       % hyperlinks
\usepackage{url}            % simple URL typesetting
\usepackage{booktabs}       % professional-quality tables
\usepackage{amsfonts}       % blackboard math symbols
\usepackage{nicefrac}       % compact symbols for 1/2, etc.
\usepackage{microtype}      % microtypography
\usepackage{lipsum}
\usepackage{fancyhdr}       % header
\usepackage{graphicx}       % graphics
\graphicspath{{media/}}     % organize your images and other figures under media/ folder
\usepackage{float}
\usepackage{caption}
\usepackage{subcaption}
\usepackage{adjustbox}
\usepackage{algpseudocode}
\usepackage{algorithm}
\usepackage{comment}
\usepackage{amsmath} 

\title{Biologically Inspired Deep Residual Networks for
	Computer Vision Applications
\thanks{\textit{\underline{Citation}}: 
\textbf{Authors. Title. Pages.... DOI:000000/11111.}} 
}

\author{
  Prathibha Varghese \\
  Research Scholar, \\
  Electronics \& Communication Department, \\
  Noorul Islam Centre for Higher Education,  \\
  Tamil Nadu, India\\
  \texttt{prathibha@sngce.ac.in} \\
   \And
  Dr. G. Arockia Selva Saroja \\
  Associate Professor, \\
  Department Electronics \& Communication Engineering, \\
  Noorul Islam Centre for Higher Education,  \\
  Tamil Nadu, India\\
  \texttt{arockiaselvasaroja@niuniv.com} \\
}

\begin{document}
\maketitle

\begin{abstract}
Deep neural network has been ensured as a key technology in the field of many challenging and vigorously researched computer vision tasks. Furthermore, classical ResNet is thought to be a state-of-the-art convolutional neural network (CNN) and was observed to capture features which can have good generalization ability. In this work, we propose a biologically inspired deep residual neural network where the hexagonal convolutions are introduced along the skip connections. The performance of different ResNet variants using square and hexagonal convolution are evaluated with the competitive training strategy mentioned  by  \cite{heresnet}. We show that the proposed approach advances the baseline image classification accuracy of vanilla ResNet architectures on CIFAR-10 and the same was observed over multiple subsets of the ImageNet 2012 dataset. We observed an average improvement by 1.35\%  and 0.48\% on baseline top-1 accuracies for ImageNet 2012 and CIFAR-10, respectively. The proposed biologically inspired deep residual networks were observed to have improved generalized performance and this could be a potential research direction to improve the discriminative ability of state-of-the-art image classification networks.   
\end{abstract}

% keywords can be removed

\section{Introduction}
\label{sec:introduction}
Deep convolutional neural networks have made significant progress in the field of computer vision \cite{imagenet_cnn, zeiler_visual, alom2018history}.  The non-linearities in high dimensional image data can be better captured with increased depth of networks. However, network performance deteriorates with increased depth of networks due to the following reasons:  
\begin{itemize}
	\item \textbf {Vanishing gradient}: Using chain rule in backpropagation, gradients (partial derivatives) are calculated. During the training process, depending upon the activation function, gradients decrease exponentially ie., the gradient of the network become smaller and smaller \cite{bengiograd, xavier_grad}.
	\item \textbf {Harder optimization}: Adding more layers to the neural networks increases more parameters, thereby leading to  increased training errors \cite{harder_opt}.
\end{itemize}
The residual network (ResNet) \cite{heresnet} architecture explicitly addresses the problem of vanishing gradients by including multiple skip connections along with the main convolution layers. These connections are contrary to the regular convolution layers and allow for easy backpropagation of the gradients without any decay. Also, due to the inherent advantages of skip-connections and its generalisation ability, ResNet is widely used as backbone feature extractor in multiple computer vision applications \cite{wu2019wider, farooq2020covid}. 

\begin{figure}[H]
	\centering
	\begin{tabular}{cc}
		\includegraphics[width=0.3\linewidth, height=0.2\textheight]{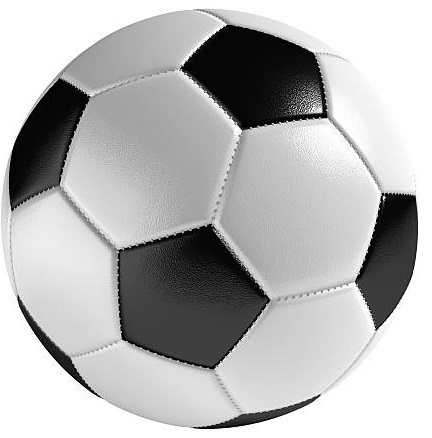} & 
		\includegraphics[width=0.3\linewidth, height=0.2\textheight]{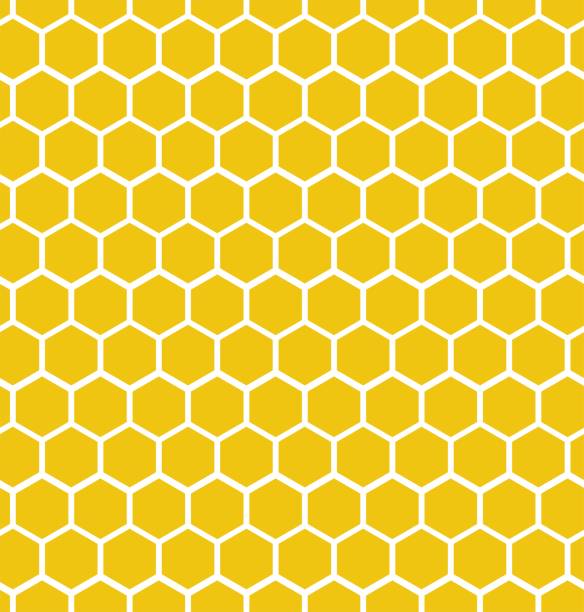} \\
		(a) & (b)     
	\end{tabular}
	\caption{Natural hexagonal structures \cite{heximages_online}. (a) Football, and (b) Honeycomb.}
	\label{fig:1_a}
\end{figure}

\begin{figure}[H]
	\centering
	\begin{tabular}{cc}
		\includegraphics[width=0.4\linewidth, height=0.2\textheight]{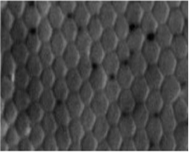} &
		\includegraphics[width=0.4\linewidth, height=0.2\textheight]{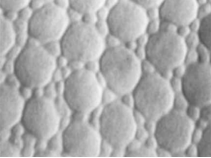} \\
		(a) & (b) 
	\end{tabular}
	\caption{Optical section arrangement of the foveal cone mosaic \cite{curcio1990human}. (a) Fovea central and (b) Fovea slope.}
	\label{fig:2_b}
\end{figure}

Majority of the neural network computations are performed on Euclidean data. However, there has been a growing interest in geometry perception of relational data and usage of those insights to learn from good non-Euclidean representations. As shown in Fig. \ref{fig:1_a}, hexagonal patterns are very commonly observed in nature. Along with football and honeycomb structures, there are few others such as bees compound eye, bubble raft, pencil base etc. Most importantly, structure of human fovea also resembles hexagonal tessellation. Fig. \ref{fig:2_b} shows the photoreceptor mosaic in the fovea captured by Nomarski differential interference contrast microscopy (NDIC) at different levels of inner segments throughout the retina \cite{curcio1990human}. Fig \ref{fig:2_b} (a) shows the profiles of rods while the Fig. \ref{fig:2_b} (b) shows the profiles of foveal cones with small intervening profiles of rods which are three times smaller than cones at retinal locations. 

On the back of the eye, the retina is a thin membrane (tissue) that aims to capture light data and transforms them as neural signals which is further sent to the brain for visual recognition. As shown in Fig. \ref{fig:2_b}, the Fovea is a small retinal area. It's structural architecture is of a highly densely conical shape and placed in a hexagonal position which captures sharper vision than the square grating layout \cite{middleton,sheridan}. Adding to that \cite{Mersereau} has demonstrated that there is an increase of 13.4$\%$ in transformation efficiency for the hexagonal pixels when compared to pixels arranged in square tessellation. This implies that hexagonal sampling requires 13.4$\%$ fewer pixels than when compared to square tessellation to capture same amount of data. This results in fewer storage requirements and less computational time. \cite{middleton} has shown that the hexagonal tessellation captures more information with identical number of sampling points when compared to the data captured on square tessellation. Apart from being close to human vision, as illustrated in \cite{Mersereau, mostafa, middleton, sheridan}, the structural arrangement of hexagonal lattice provides added advantage over the square lattice counterpart. The authors in \cite{middleton} and \cite{sheridan} show that there is an improved precision on the detection of circular and near circular edges of hexagonal images. 

Any attempts in building the imaging systems whose sensing mechanism and data processing abilities mimic the way human visual system captures and processes the data will aid in building better systems close to human level performance. However, due to the ease of practical realisation of mathematical operations defined on Euclidean space, data captured on square tessellations are more popular. Hex-mov predictions \cite{Neurohex}, imaging atmopsheric Cherenkov telescope (IACT) \cite{iact}, and ice cube data analysis \cite{icecube} are some of the works which deal with the data captured by sensors arranged on hexagonal lattice. 

In this paper, we propose a biologically inspired hybrid architecture Hex-ResNet to improve the generalisation ability of deep residual networks through hexagonal convolutional filters. We show that the baseline performance of ResNet architecture can be increased by incorporating minimal hexagonal filters along a few of the skip connections. The key highlight of our approach is to effectively combine the advantages offered by each of the individual (square and hexagonal) tessellations. Unlike \cite{hoogeboom2018hexaconv} which operate directly on a hexagonal lattice, we use the implementation of \cite{hexagdly} which implements efficient routines for hexagonal convolution by using an equivalent sum of two square convolutions. This way, we obtain the gain in performance by using the efficient computational support defined on square tessellations. Our proposed architecture has achieved improved validation and testing accuracy as compared with the several existing ResNet models in terms of Top-1 and  Top-5 accuracy. We also show that this enhanced performance was obtained without significant increase in computational overhead. We summarise our contributions and findings as follows:

\begin{itemize}

	\item 	We modified the vanilla ResNet by incorporating hex convolutions along a few of the skip connections and found an improvement in the baseline image classification accuracy.

	\item 	We validated the efficiency of the newly proposed Hex-ResNet architecture, by performing extensive experiments on CIFAR-10 and ImageNet 2012 benchmark datasets.
	
	\item   We show that  incorporating hex convolutions along the skip connection paths, consistently improves the baseline image classification accuracy across multiple ResNet configurations.     
	
\end{itemize}

The remainder of this paper is organized as follows: a review of the related works is given in section 2 which covers all the related works to our approach. Section 3 initially presents the preliminaries of classical ResNet architectures, skip connections followed by the details of our proposed Hex-ResNet architecture. In Section 4, we give the details of experimental results and training procedures used on our proposed architecture using CIFAR-10 and ImageNet datasets. We finally conclude with our observations in section 6.

\section{RELATED WORKS}
\label{sec:related_works}

\subsection{Residual Networks}

The ability of deep residual networks to overcome the vanishing gradient problem through skip connections made it an ideal candidate as a backbone feature extractor for various computer vision tasks \cite{heresnet,farooq2020covid,resnetapp1,resnetapp2}. \cite{baoqi} proposed an improved ResNet via adjustable shortcut connections, and designed a convex k strategy using different region parameters changing rules. \cite{resnet_timm} shared competitive training settings and pre-trained models for several ResNet configurations in the Timm open-source library. Pre-activation ResNets were introduced in the study \cite{identity_mappings} by rearranging the elements in the building block to improve the signal propagation path. All the popular works which used ResNet as a backbone feature extractor operated on square lattice.

\subsection{Hexagonal Processing}
\label{sec:hex_processing}
Data stored on hexagonal lattice was shown to have significant advantage over the square lattice \cite{hexadv1, hexadv2, hexadv3}. However, most of the existing display devices capture data using pixels arranged in square tessellation due to the ease of realising many mathematical operations defined on Euclidean geometry. This current generation devices which capture data on square lattice will not have the ability to take advantage of multiple benefits offered by hexagonal lattice such as equidistance, higher
angular resolution, smaller quantization error, greater circular symmetry, and efficient sampling scheme \cite{hexadv4, hexadv5, hexadv6}. Also, as shown in Fig. \ref{fig:2_b}, the imaging sensors in human visual system are in the form of hexagonal tessellation. Hence, for any intelligent visual system to have human level performance they need to replicate the way data is captured and processed in human visual system. 

There has been a growing interest in building biologically inspired models in the recent past. B-spline models have been successfully used for tasks such as interpolation and sampling on square lattices. \cite{van2004hex} has introduced hex-splines for hexagonal sampled data. \cite{middleton2001edge} has reported an improved edge detection performance for curved objects sampled on hexagonal lattice. \cite{abbas2015pet} has shown an improvised PET image reconstruction on hexagonal lattice using a filtered back projection and sparse dictionary based denoising approach. The authors in \cite{contreras2014hexagonal, ortiz2011hexagonal} proposed an improvised image enhancement technique for hexagonal resampled ultrasound images. 
%A suitable addressing scheme will be useful to access pixels on hexagonal lattice. Fig \ref{fig:addressing} shows a few of the addressing schemes used in the literature \cite{addressing}.   \textbf{Some details about the addressing schemes would be useful here.} 
While many existing approaches convert the data defined on square lattice to hexagonal tessellation, there are a few approaches \cite{hexdev1, hexdev2, hexdev3} which constructed imaging and display devices which directly captures data on a hexagonal lattice. 

\subsection{Hexagonal Convolutional Neural Networks}
\label{sec:hex_networks}
\cite{zhao2020hexcnn} has proposed a deep learning based approach which performs both forward and backward directly on hexagonal data as input and thus overcoming the limitation of ZeroOut which tend to add significant computational overhead.  \cite{hoogeboom2018hexaconv} has extended the theory of group equivariant convolutional neural networks to hexagonal lattice so as to exploit the six-fold symmetries in $p6$ and $p6m$ groups. There are few other works \cite{hexdnn1, hexdnn2} which utilise the square convolutional routines for hexagonal lattices. Unlike the previous techniques, the proposed approach is a hybrid technique which use the square equivalent of hexagonal filters along the skip connections of deep residual networks and show an improvised performance over the classical ResNet architectures. Our approach is an attempt to exploit the advantages offered by both square as well as hexagonal tessellations. Also, we make use of existing square convolutional routines offered by popular deep learning frameworks like PyTorch to perform equivalent hexagonal convolutions. 

\begin{comment}
	\begin{figure}[!h]
		\centering
		\includegraphics[width=0.96\linewidth, height=0.25\textheight]{4new}
		\caption{Different coordinate systems a) offset b) doubled c) axial d) cube.}
		\label{fig:addressing}
	\end{figure}
\end{comment}

\section{Proposed approach}
\label{sec:preliminaries}
In this section, we will initially describe the preliminaries followed by the details of proposed approach. 
\subsection{Classical Residual Network Architecture}
\label{sec:resnet_summary}
The architecture of various configurations of classical ResNet \cite{heresnet} is explained in Table \ref{fig:t2}. Initially, all the images were convolved to 64 layers using a 7$\times$7 convolution with stride 2.   Instead of applying shortcut connections to a pair of 2$\times$2 filters,  they were applied to the pair of 3$\times$3 filters.    For relatively small networks like 18 and 34 layers non bottlenecked blocks were used which consisted of 4 convolution blocks each having two 3$\times$3 convolutions.   For ResNet-18, these convolutions were repeated twice for each filter map \{64,  128,  256,  512\} and for 34-layer ResNet they were repeated in the order \{3,  4,  6,  3\}.    To make ResNet more economical on large datasets such as ImageNet, a deep bottleneck architecture was used with 50 and 101 layers.  Bottleneck architecture contains a stack of 1$\times$1 conv,  3$\times$3 Conv and a 1$\times$1 Conv repeated.  In the last 1$\times$1 convolutions, the output channels are increased four times.   For ResNet 50, these blocks are repeated \{3,  4,  6,  3\} times and for ResNet 101 these blocks are repeated as \{3,  4,  23,  3\}.  Option B shortcut is used for increasing dimensions.   After the blocks, an average pool,  fully connected layer and, softmax were also applied.  

\begin{table}[H]
	\centering
	\caption{ResNet architectures for ImageNet/CIFAR-10 dataset.}
	\includegraphics[width=0.55\linewidth, height=0.22\textheight]{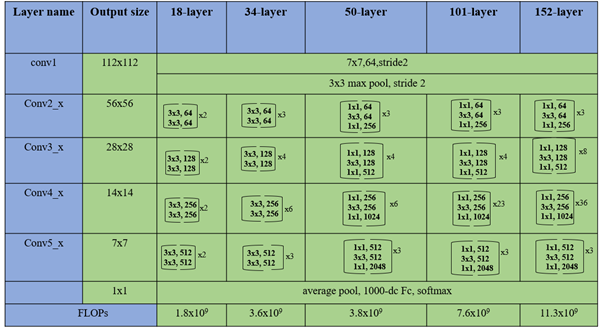}
	\label{fig:t2}
\end{table}

\subsection{Skip Connections}
\label{sec:skip_connections}
The residual shortcuts introduced in \cite{heresnet} are meant for the easy back propagation of gradients and hence should not introduce any significant computational overload. Based on the input and output dimensions of image/features, the residual shortcuts are classified into two categories: 

\begin{figure}[H]
	\centering
	\begin{tabular}{cc}
		\includegraphics[width=0.26\linewidth, height=0.18\textheight]{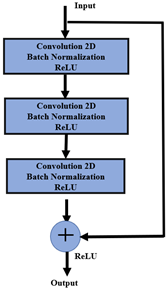} &
		\includegraphics[width=0.26\linewidth, height=0.18\textheight]{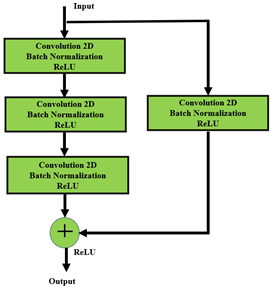} \\
		(a) & (b) 
	\end{tabular}
	\caption{Fundamental block of ResNet architecture with skip connections. (a) Identity shortcut \cite{heresnet} and (b) Projection shortcut \cite{chen2020deep}.}
	\label{fig:3_a}
\end{figure}

\begin{itemize}
	\item \textbf{Identity shortcut}:
	Identity shortcut mapping can be used, when the input and output dimensions match, it just requires simple mathematical additions for tensors and can be utilised directly. It is really hard to find a direct mapping from input to output using a standard neural network block.  So, instead of that,  the difference between the input and output vectors is calculated using the residual block. As in \cite{improved_resnet}, identity shortcut is used along with convolution, batch normalisation, max pooling, and activation function as shown in Fig. \ref{fig:3_a} (a). and, fed as input to the add operator. Thus, having a direct mapping plus the input passed to the activation function.

	\item \textbf{Projection shortcut}:
	The projection shortcuts (as shown in Fig. \ref{fig:3_a} (b)) can be used to capture the information which has not be derived or extracted along the main information path in the residual ResNet block. This include $1\times 1$ convolution, batch normalization, and ReLU units. Linear projection shortcut are also preferred when the input and output dimensions of residual ResNet block are non-identical. The output of this projection shortcut is combined with the output of fundamental block through the addition operator as shown in Fig. \ref{fig:3_a} (b). 
\end{itemize}

In the proposed approach, we will replace the square convolutions with hexagonal convolutions along the projection shortcuts of residual ResNet blocks Fig. \ref{fig:3_a}. However, we leave the rest of identity shortcuts without any modifications. The proposed approach to modify a few of the projection shortcuts will not add any significant computational overhead and thus avoiding the vanishing gradient problem. Through our experimental results, we will show that our approach can aid in extracting better representations over the classical ResNet architectures. In the next subsection, we will describe the details of hexagonal convolutions. 

\subsection{Hexagonal Convolution Operations}
\label{sec:hexagdly}
\cite{hexagdly} proposed HexagDLy, a pytorch package which performs hexgonal convolutions and pooling by blending multiple rectangle-shaped filters with different dilations and sizes. They define the size of kernel based on the number of levels of neighbouring elements. These hexagonal convolutions achieve higher efficiency than ZeroOut method \cite{zhao2020hexcnn} due to its divide-and-conquer approach. 

\begin{figure}[H]
	\centering
	\begin{tabular}{cc}
		\includegraphics[width=0.2\linewidth,height=0.15\textheight]{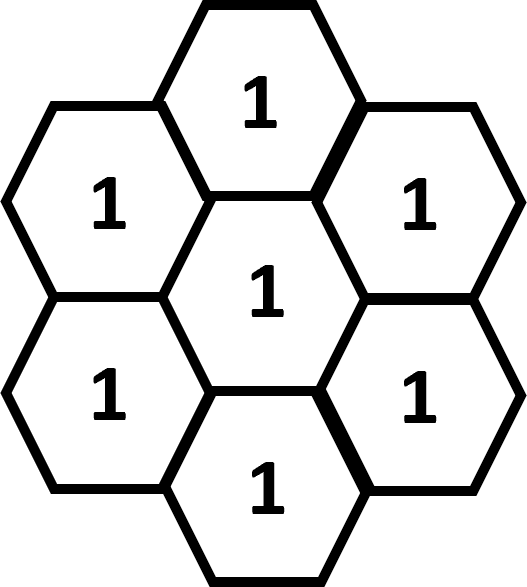} &  
		\includegraphics[width=0.2\linewidth, height=0.15\textheight]{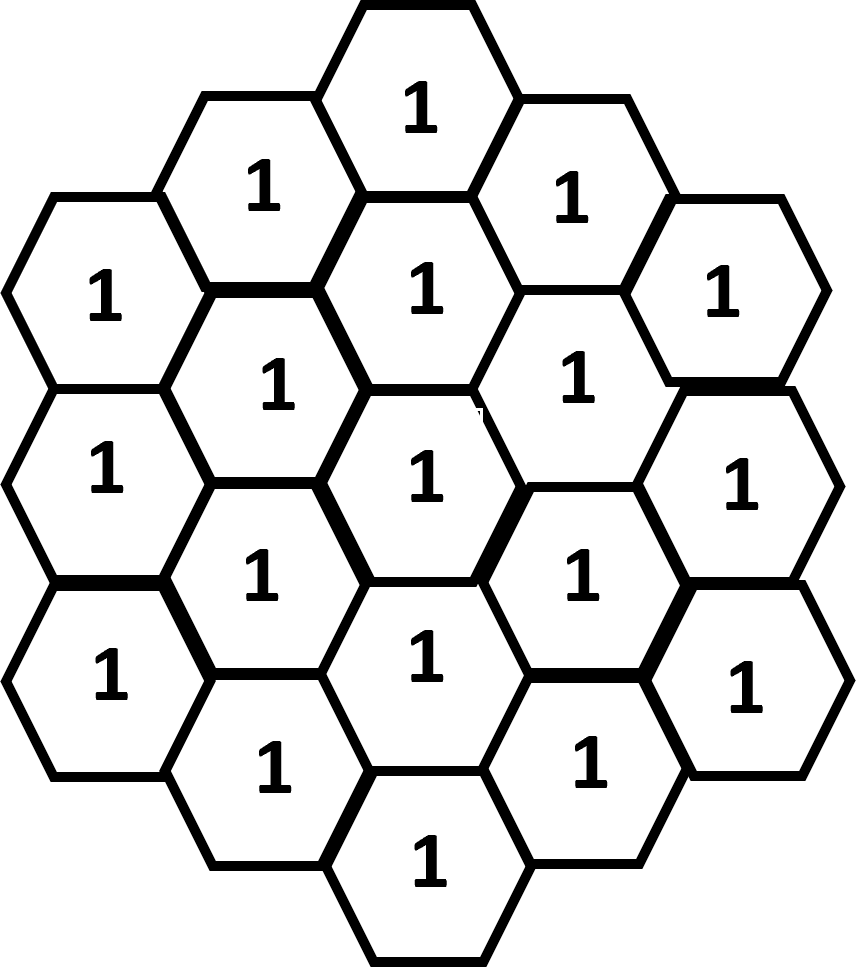}\\
		(a) & (b)
	\end{tabular}
	\caption{Size of hexagonal kernels. (a) Size 1 and (b) Size 2.}
	\label{fig:4_a}
\end{figure}

Increasing the number of learnable parameters along the skip connection of residual network architecture will again lead to vanishing gradient problem as mentioned in Sec. \ref{sec:introduction} and hence, we will limit to use size one hexagonal kernels in our approach. Figs. \ref{fig:4_a} (a) and (b) shows sample hexgonal kernels of size one and two, respectively. Following \cite{hexagdly}, we will use the scheme so as to avoid any significant overhead involved in the interpolation computations while converting from square to hexagonal tessellations or vice-versa. 

\begin{figure}[H]
	\centering
	\includegraphics[width=0.5\linewidth, height=0.2\textheight]{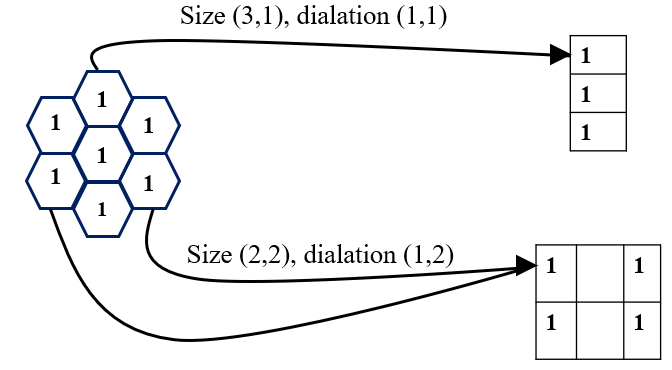}
	\caption{Split of size one hex kernel into equivalent rectangular kernels.}
	\label{fig:5}
\end{figure}

\begin{figure}[H]
	\centering
	\includegraphics[width=0.6\linewidth, height=0.3\textheight]{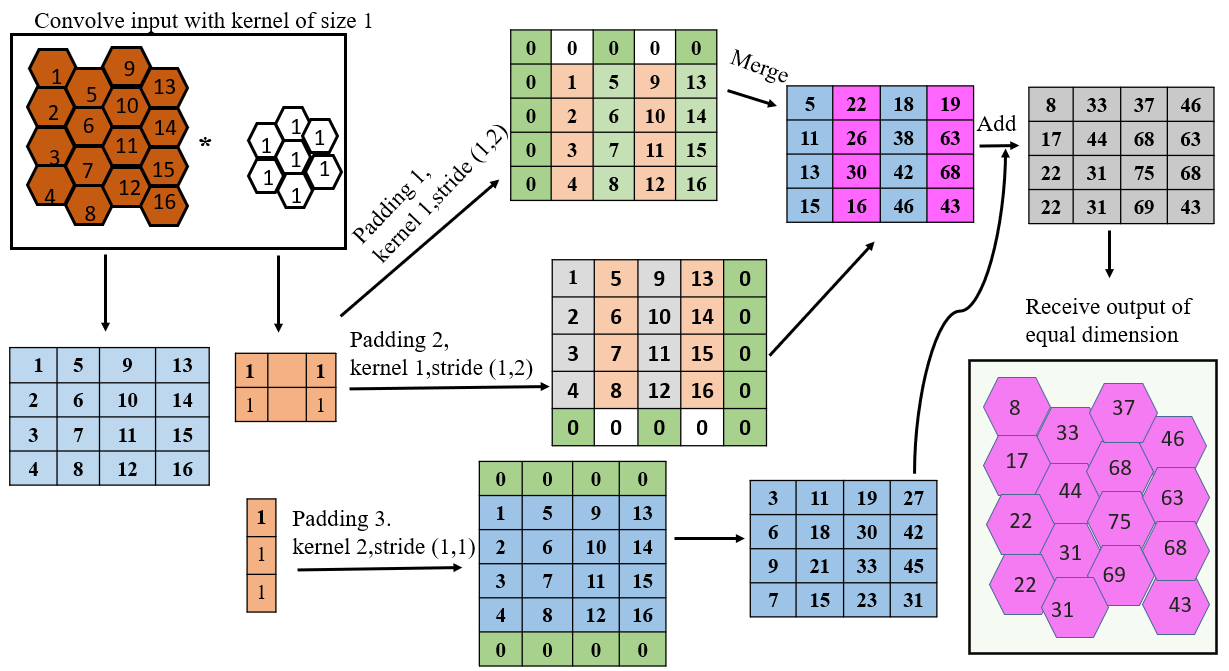}
	\caption{Implementation of hexagonal convolution \cite{hexagdly}}
	\label{fig:6}
\end{figure}

Fig. \ref{fig:5} shows the an example of hexagonal convolution using a hexagonal kernel of size one. Let $\mathbf{H}$ be the image data defined on hexagonal tessellation and $\mathbf{S}$ be its equivalent image defined on square tessellations. We denote the hexagonal kernels by $\mathbf{K}^l$, where $l$ indicates its size. For the ease of mathematical demonstration, we have used the kernel weights as unity. But in the final implementation, we use the trainable kernel weights. As shown in Fig. \ref{fig:6}, we derive equivalent rectangular kernels ($\mathbf{K}^1_{r1} \in \mathbb{R}^{2 \times 3}$ and $\mathbf{K}^1_{r2} \in \mathbb{R}^{3 \times 1}$) corresponding to $\mathbf{K}^1$. Note that a hexagonal kernel of size one will have two equivalent rectangular kernels. Similarly, a hexagonal kernel of size $l$ will have $l+1$ equivalent rectangular kernels. The convolutions with the kernels $\mathbf{K}^1_{r1}$ and $\mathbf{K}^1_{r2}$ can be now easily done with efficient routines available in PyTorch. However, in order to obtain hexagonal convolutions through rectangular kernels, we need to appropriately pad $\mathbf{S}$ in three different ways as shown in Fig. \ref{fig:6}. Let $\mathbf{S}_1$, $\mathbf{S}_2$, $\mathbf{S}_3$ be the three different padded versions of $\mathbf{S}$. Mathematically, the convolution of rectangular kernels $\mathbf{K}^1_{r1}$ and $\mathbf{K}^1_{r2}$ with $\mathbf{S}_1$, $\mathbf{S}_2$, $\mathbf{S}_3$ can be represented as 
\begin{align}
	\mathbf{P}_1 &= \mathbf{S}_1 \ast_{(1,2)} \mathbf{K}^1_{r1} \\
	\mathbf{P}_2 &= \mathbf{S}_2 \ast_{(1,2)} \mathbf{K}^1_{r1} \\
	\mathbf{P}_3 &= \mathbf{S}_3 \ast_{(1,1)} \mathbf{K}^1_{r2} \\ \nonumber
\end{align}
where $\mathbf{P}_1$, $\mathbf{P}_2$, $\mathbf{P}_3$ denote the result of convolution with the kernels $\mathbf{K}^1_{r1}$ and $\mathbf{K}^1_{r2}$. The operator $\ast_{(x,y)}$ denote the convolution with stride of $x$ and $y$ units along the rows and columns, respectively. The next step is to merge $\mathbf{P}_1$ and $\mathbf{P}_2$ by picking the alternate columns as shown in Fig. \ref{fig:6}. Mathematically we denote the merge operation as follows
\begin{equation}
	\mathbf{P}_{12} = MERGE\left( \mathbf{P}_1, \mathbf{P_2} \right).
\end{equation}
The square equivalent of hexagonal convolution is obtained by one final addition operation as:
\begin{equation}
	\mathbf{Q} = \mathbf{P}_{12} \oplus \mathbf{P}_3
\end{equation}
where $\oplus$ denotes the element-wise addition operation. The output $\mathbf{Q}$ is rearranged to hexagonal lattice if needed for further processing as shown in Fig. \ref{fig:6}.

\subsection{Hex-ResNet}
\label{sec:Hex_resnet}

\begin{figure}[H]
	\centering
	\includegraphics[width=0.5\linewidth, height=0.25\textheight]{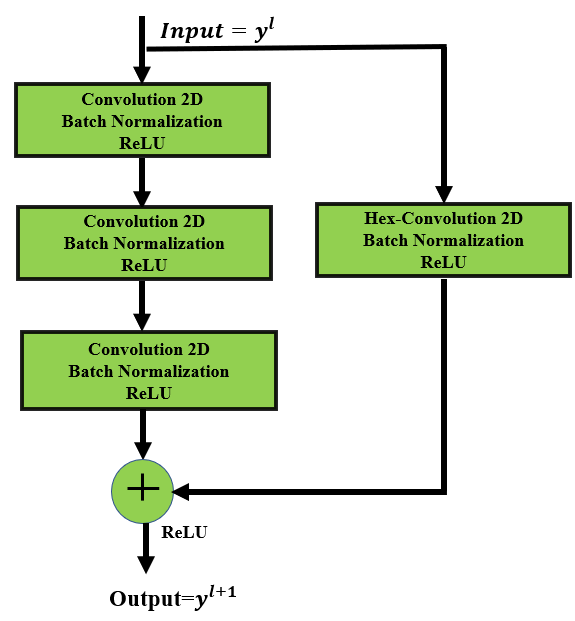}
	\caption{Fundamental block of Hex-ResNet.}
	\label{fig:7}
\end{figure}

\begin{figure}[H]
	\centering
	\includegraphics[width=0.2\linewidth, height=0.9\textheight]{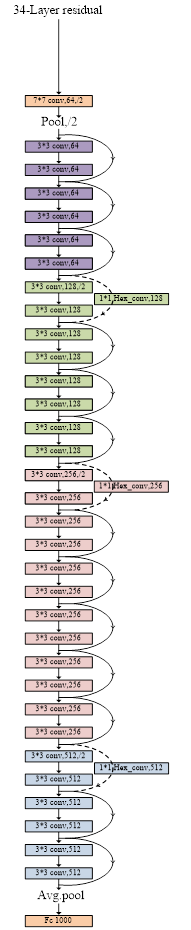}
	\caption{Hex-Residual Network Architecture for ImageNet and CIFAR-10. (Readers are requested to zoom in to view the details.)}
	\label{fig:8}
\end{figure}

In this section, we describe the details of proposed Hex-ResNet architecture. Our model is a hybrid approach where we combine the symmetries offered by both the square and hexagonal tessellations. The residual block of the proposed architecture is shown in Fig. \ref{fig:7} where we incorporate hexagonal convolutions along with batch normalisation (BN) (used for training convergence) and rectified linear unit introduces the non-linearity into the network. We employ regular $3 \times 3$ convolution in the normal path with padding and stride set to one. In the projection shortcut path, 2D Hex-convolution  with kernel size set to one is used. In order to match the input and output vectors of this residual block, a variable stride is used with padding set to zero. Each residual block can be mathematically modelled as Eq. \ref{eq:key1}. 

\begin{equation}
	y^{\left[l+1\right]}=\begin{cases}ReLU(F(y^{\left[l\right]},
		\left\{P_i^{\left[l+1\right]}\right\})+y^{\left[l+1\right]}),\\ 
		\left(if\:size\left(F(y^{\left[l\right]},\left\{P_i^{\left[l\right]}\right\})\right)=size(y^{\left[l\right]})\right);\\ ReLU(F(y^{\left[l\right]},
		\left\{P_i^{\left[l+1\right]}\right\})+P_p^{\left[l\right]}y^{\left[l\right]}),\\\left(if\:size\left(F(y^{\left[l\right]},\left\{P_i^{\left[l\right]}\right\})\right)\neq size(y^{\left[l\right]})\right);\end{cases}
	\label{eq:key1}
\end{equation}

where $ y^{\left[l\right]} $ and $ y^{\left[l+1\right]} $ are the input and output vectors of the $l^{th} $ Residual block. F($ y^{\left[l\right]},\{P_i^{\left[l\right]}\} $)  is a learnable residual mapping function that can have several layers. $ P_p^{\left[l\right]} $ is a learnable linear projection matrix that allows mapping the size of $ y^{\left[l\right]} $ to the output size of F, which exists only in case of not corresponding dimensions for performing the element-wise addition between F and $y^{\left[l\right]}$. In our approach, since we combine the hexagonal (along projection shortcut) and square convolutions (main path of residual block) we use $\mathbf{Q}$ as it is for further processing without mapping it to hexagonal lattice. Note that we limit to use size one hexagonal kernel so as to avoid vanishing gradient problem. Hexagonal convolution is performed using the procedure described in Fig. \ref{fig:8}. The proposed Hex-ResNet framework with a modified projection shortcut is shown in Fig. \ref{fig:8} where the total number of weighted layers are 34. It is modelled by stacking together many residual blocks. The final layer includes a global average pooling layer and a M-fully connected layer with softmax activation function, where M denotes the number of classes in the dataset. The same is followed for realising other ResNet configurations with multiple layers as well. Note that we replace only the convolutions along the projection shortcuts but we retain the rest of skip connections without any modifications. The reason behind retaining the rest of skip connections without any modification is to avoid the vanishing gradient problem.

\section{Experimental Results}
\label{sec:exp_results}
In this section, we describe the details of our experimental results on CIFAR10. Also, we demonstrate the effectiveness of our approach on a subset of ImageNet. Following \cite{heresnet}, we use Top-1 and Top-5 accuracy as the performance metric.

\subsection{Software}
We use \cite{hexagdly}, a PyTorch based library which has routines for computing hexagonal convolutions. The combination of precise padding and striding scheme is used to preserve the hexagonal structure of the input image at the output side.    
\subsection{Image Input Format}
The hexagonal grid can be understood as two stacked rectangular grid structures on a distinct square array. By shifting rows and columns half the distance between consecutive nearby pixels, proper horizontal and vertical alignment is achieved, resulting in a square grid array with rows and columns aligned as shown in Fig. \ref{fig:6}. 
\subsection{Hardware}
Google colab pro+ is used which is a web application platform. This has many built-in package that enables us to train our model on high-end GPUs. The GPU device that was allocated during my experiments was mostly Tesla V100-SXM2-16GB with compute capability 7.0" and A100-SXM4-40GB with compute capability 8.0". 
\subsection{Datasets}
\subsubsection{CIFAR10}
\begin{figure}[H]
	\centering
	\begin{tabular}{cc}
		\includegraphics[width=0.45\linewidth, height=0.25\textheight]{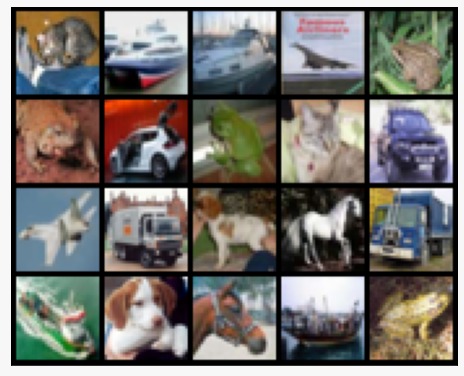} & 
		\includegraphics[width=0.45\linewidth, height=0.25\textheight]{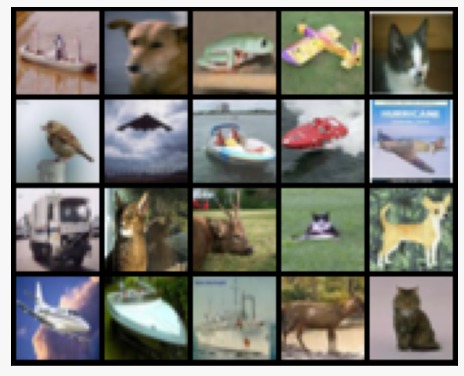} \\
		(a)   & (b)
	\end{tabular}
	\caption{Sample images from CIFAR10: a) Training dataset and b) Testing dataset.}
	\label{fig:9_a}
\end{figure}

The CIFAR-10 \cite{cifar} is one of the benchmark datasets used for image classification. This dataset  has 60000 images sorted into 10 different groups and is used to test the proposed architecture. The entire dataset is divided into training set with 50,000 images and a test set with 10,000 images. All the training images were padded with 4 pixels on each side and a $32\times 32$ crop was randomly sampled from the padded image or its horizontal flip. No augmentations were applied to the test dataset. The augmented train dataset was then split into train and validation containing 45,000 and 5000 images respectively. Fig. \ref{fig:9_a} shows the sample images of the training and test images of CIFAR-10 dataset from different classes. Cross entropy loss was the criterion with stochastic gradient descent as optimiser (learning rate=0.1, momentum=0.9, weight decay=0.001). The model was then trained for 182 epochs with the learning rate divided by 10 at 32k,  48k,  and 64k iterations.   

\subsubsection{ImageNet:}
\begin{figure}[H]
	\centering
	\begin{tabular}{cc}
	\includegraphics[width=0.45\linewidth, height=0.26\textheight]{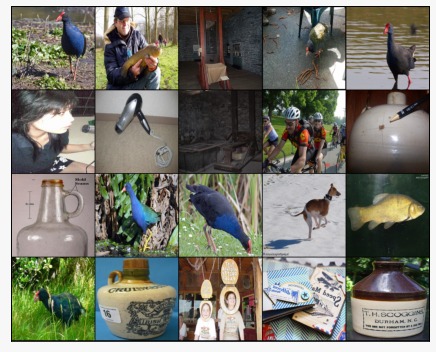} & 
		\includegraphics[width=0.45\linewidth, height=0.26\textheight]{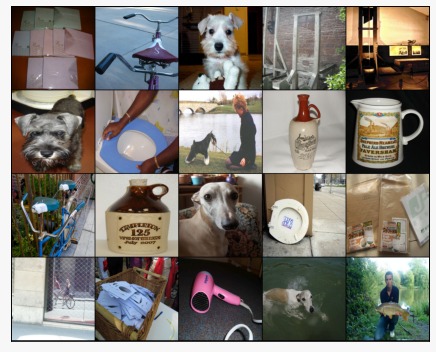} \\
		(a)   & (b)
	\end{tabular}
	\caption{Sample images from ImageNet 2012: a) Training dataset and b) Testing dataset.}
	\label{fig:10_a}
\end{figure}

We trained our Hex-ResNet architecture on ImageNet 2012 which contains more than 1.28 million images distributed among 1000 classes. Due to the constraint in the available resources, we have limited our experiments to randomly sampled 10 classes from available 1000 classes. We repeat our experiments on ImageNet on randomly sampled five different subsets, where each has 10 different classes. Fig. \ref{fig:10_a} shows the sample of images from ImageNet dataset. Each subset was further randomly split in 80:20 ratio into a train and validation sets, respectively. Each training subset contains about 4270 images and validation contains about 1000 images. We sample the same classes from test set as was used in training and validation sets. Random horizontal flip was used as the data augmentation technique and the images are resized to $224\times 224$ resolution. We use the batch of 32 images while training the model. The optimiser used is Stochastic Gradient Descent. 

\subsection{Quantitative Analysis}
\begin{figure}[H]
	\centering
	\begin{tabular}{c}
		\includegraphics[width=0.52\linewidth, height=0.22\textheight]{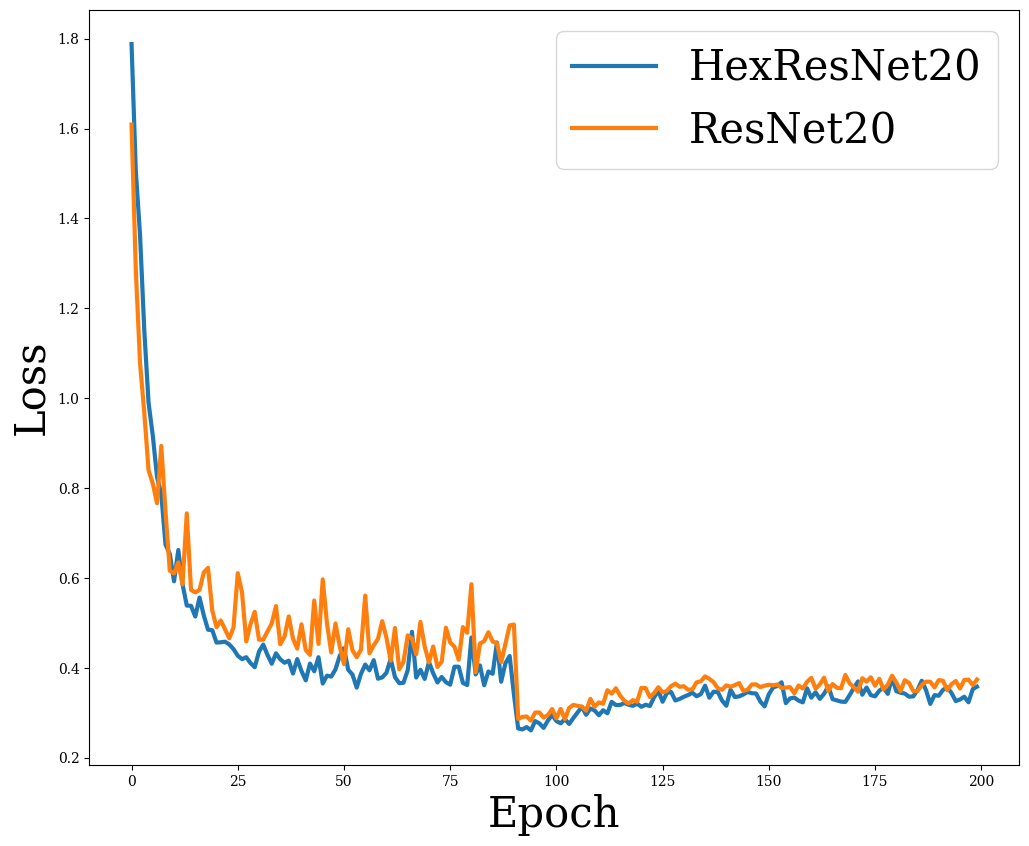}    \\
		(a)  \\
		\includegraphics[width=0.52\linewidth, height=0.22\textheight]{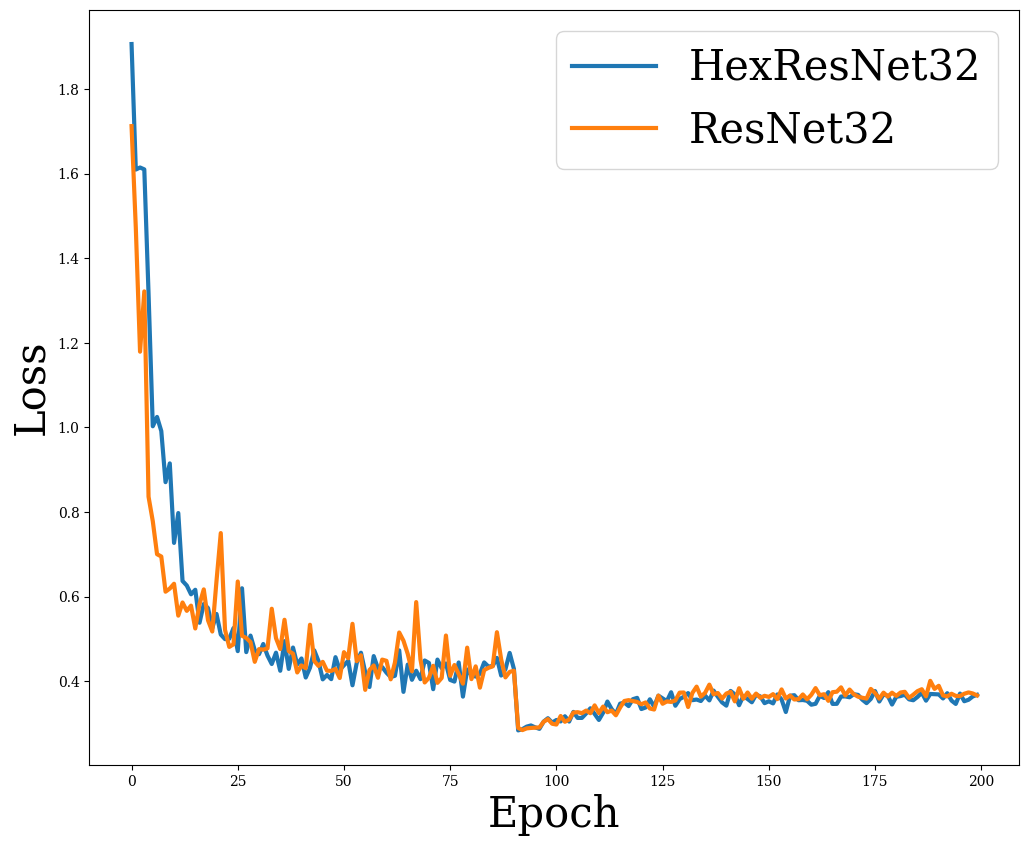}   \\
		(b)     \\
	\includegraphics[width=0.52\linewidth, height=0.22\textheight]{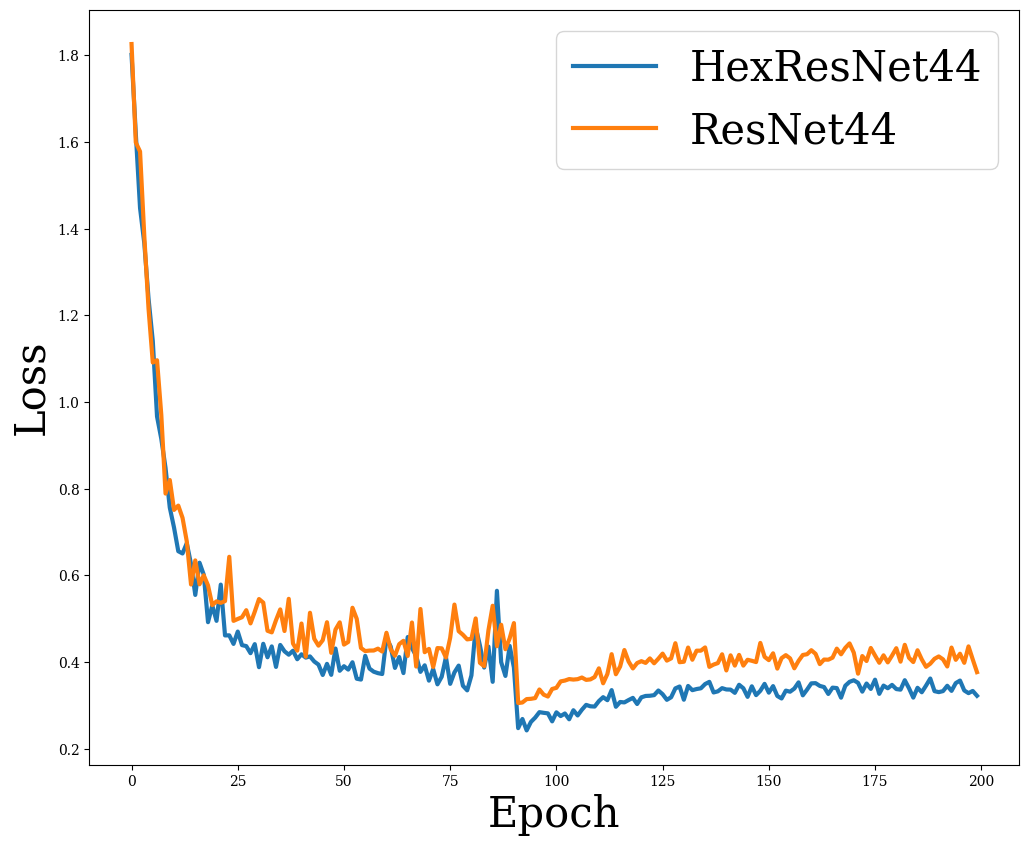}   \\
		(c)     \\
	\includegraphics[width=0.52\linewidth, height=0.22\textheight]{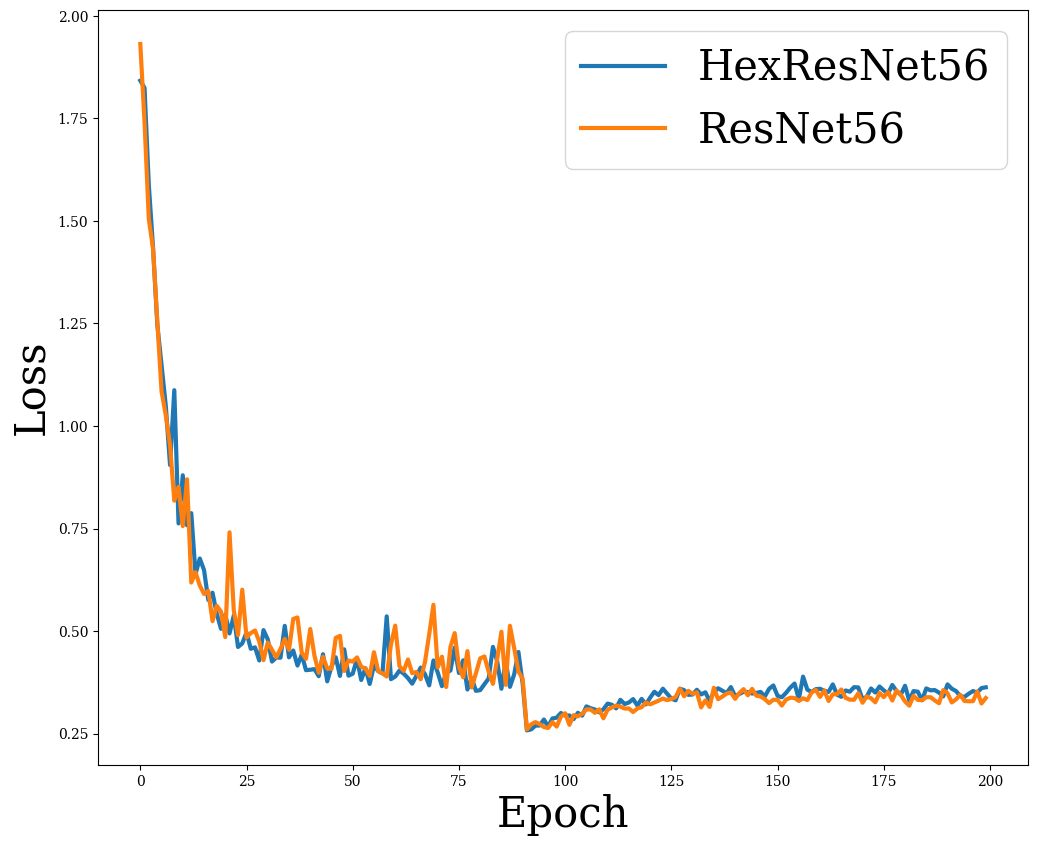}   \\
		(d)     \\
	\end{tabular}
	\caption{Validation Loss variation with respect to epochs for CIFAR-10 dataset. ResNet vs HexResNet: (a) 20 layers, (b) 32 layers, (c) 44 layers, and (d) 56 layers.}
	\label{fig:11_a}
\end{figure}

In this section, we present the quantitative analysis of the proposed approach on CIFAR-10 and ImageNet datasets. Table \ref{tab:2} shows the Top-1 and Top-5 accuracy and error percentage for CIFAR-10 for various ResNet configurations. It can be seen that the Hex-ResNet 20, 32, 44, and 56 has slightly greater validation accuracy and less Top-1 error compared to the ResNet counterparts with same number of layers, respectively. Also, Hex-ResNet brings down the top-1 error by 0.9\%  when compared to ResNet model variant 20 and by 0.2\% in higher ResNet model 56. Figs. \ref{fig:11_a} (a)-(d) depicts the validation loss vs epochs for both the base ResNet and Hex-ResNet configurations. Note that Hex-ResNet converges faster as well as has lower validation loss when compared to their ResNet counterparts. This clearly indicates the feature representation obtained through the proposed architecture has improved generalisation ability when compared to their respective ResNet configurations.

\begin{table}[H]
	\centering
	\caption{Error rates and accuracy percentage on CIFAR-10.}
	\begin{adjustbox}{width=0.8\textwidth}
		\label{tab:2}
		\begin{tabular}{|c|c|c|c|c|c|c|c|}
			\hline
			\textbf{Model} & \textbf{Parameters} & \textbf{\begin{tabular}[c]{@{}c@{}}Validation \\ Acc\end{tabular}} & \textbf{\begin{tabular}[c]{@{}c@{}}Top1 \\ acc \%\end{tabular}} & \textbf{\begin{tabular}[c]{@{}c@{}}Top 1 \\ error \%\end{tabular}} & \textbf{\begin{tabular}[c]{@{}c@{}}Top 5 \\ acc \%\end{tabular}} & \textbf{\begin{tabular}[c]{@{}c@{}}Top 5 \\ error \%\end{tabular}}   \\ \hline
			ResNet - 20 & 272, 474 & 91.  64\% & 91.  44\% & 8.  56\% & 99.  63\% & 0.  37\%   \\ \hline
			\textbf{ResNet - 20(H)} & \textbf{287, 130} & \textbf{92.    12\%} & \textbf{92.    36\%} & \textbf{7.  64\%} & \textbf{99.    67\%} & \textbf{0.  32\%}   \\ \hline
			ResNet - 32 & 466, 906 & 92.  55\% & 92.  03\% & 7.  96\% & 99.  74\% & 0.  26\%   \\ \hline
			\textbf{ResNet - 32(H)} & \textbf{481, 114} & \textbf{92.    55}\% & \textbf{92.    25}\% & \textbf{7.  75\%} & \textbf{99.    73\%} & \textbf{0.  27\%}   \\ \hline
			ResNet - 44 & 661, 338 & 91.  83\% & 92.  14\% & 7.  85\% & 99.  76\% & 0.  23\%   \\ \hline
			\textbf{ResNet - 44(H)} & \textbf{675, 098} & \textbf{92.    92\%} & \textbf{92.    44\%} & \textbf{7.  56\%} & \textbf{99.    76\% }& \textbf{0.  23\%}   \\ \hline
			ResNet - 56 & 855, 770 & 92.  97\% & 92.  16\% & 7.  84\% & 99.  69\% & 0.  31\%   \\ \hline
			\textbf{ResNet - 56(H)} & \textbf{869, 082} & \textbf{93.    14\%} & \textbf{92.    35\%} & \textbf{7.  65\%} & \textbf{99.    73\%} & \textbf{0.  27\%}   \\ \hline
		\end{tabular}
	\end{adjustbox}
\end{table}

\begin{figure}[H]
	\centering
	\begin{tabular}{c}
		\includegraphics[width=0.52\linewidth, height=0.22\textheight]{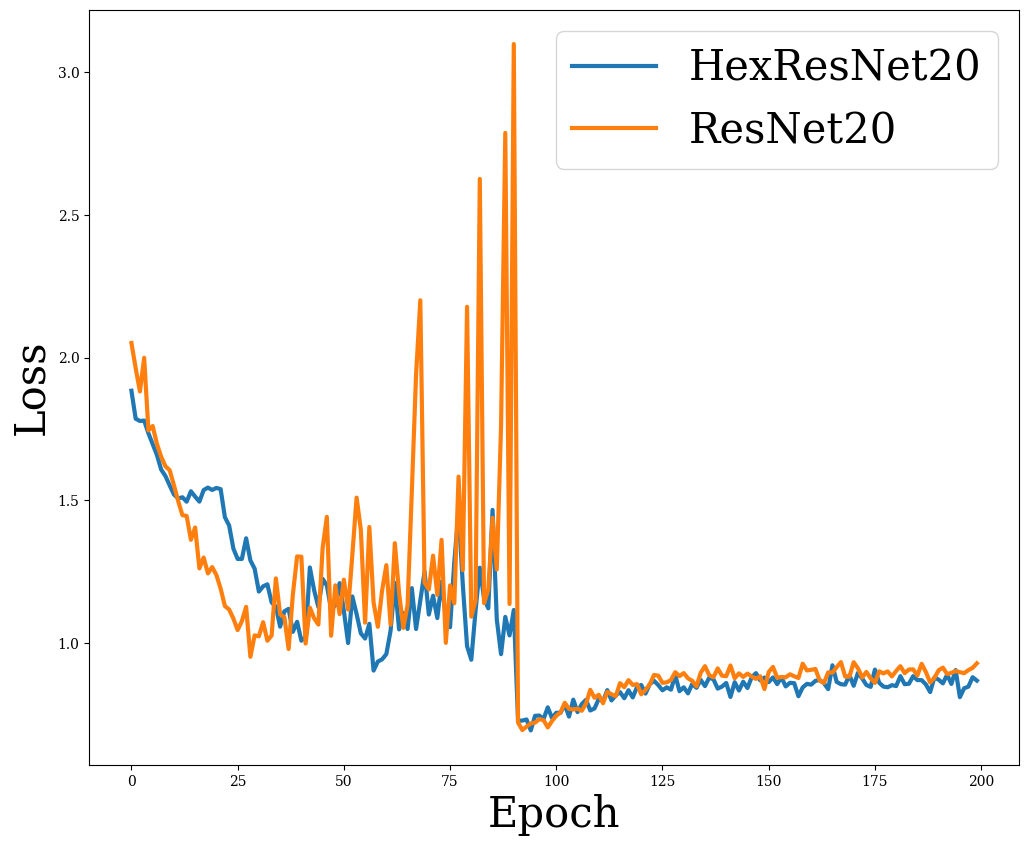}    \\
		(a)  \\
		\includegraphics[width=0.52\linewidth, height=0.22\textheight]{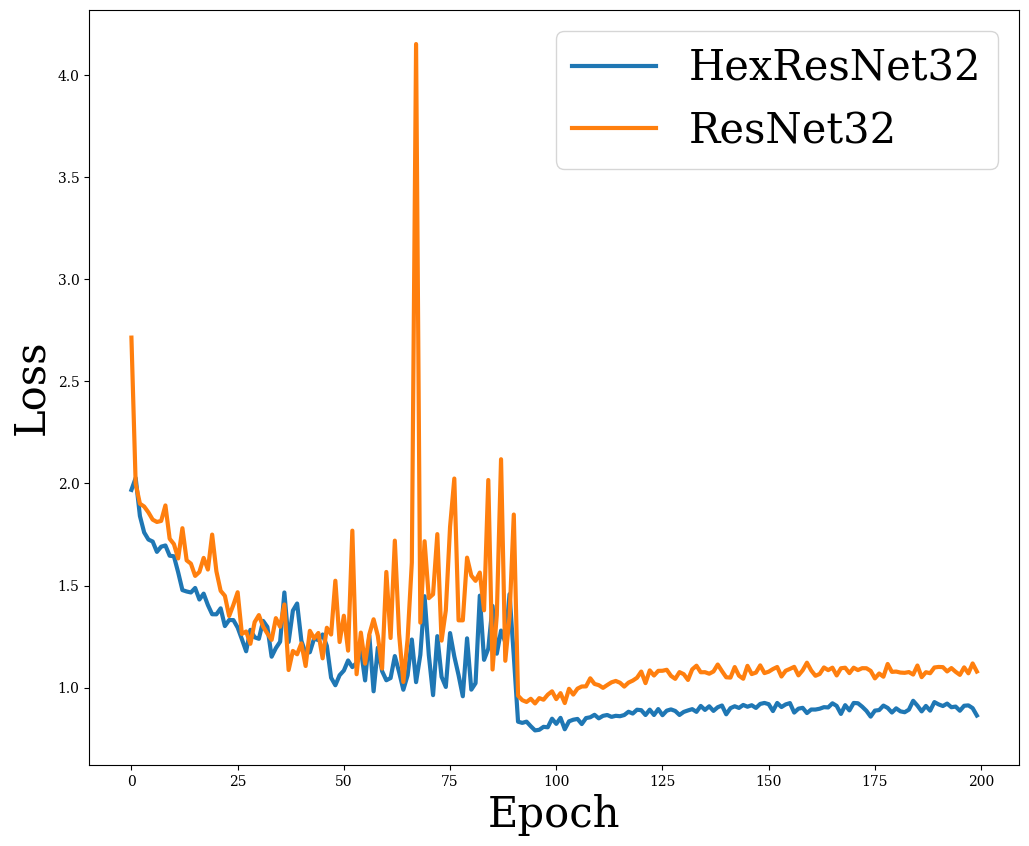}   \\
		(b)     \\
		\includegraphics[width=0.52\linewidth, height=0.22\textheight]{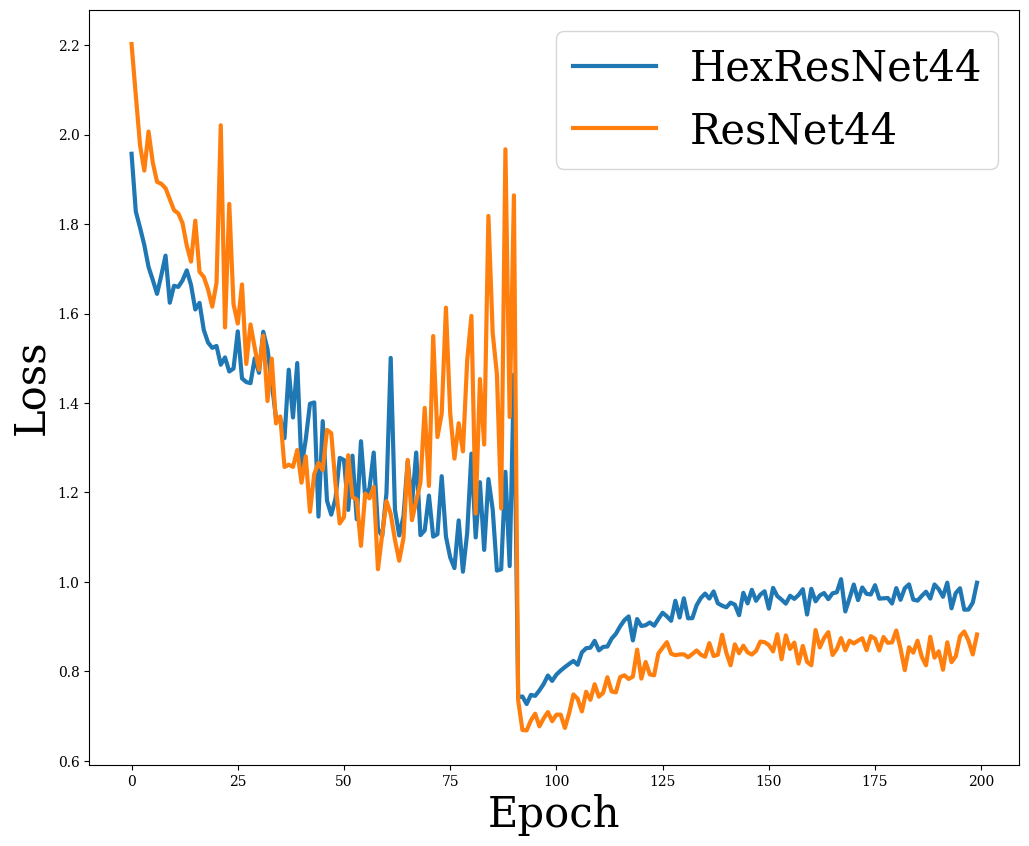}   \\
		(c)     \\
		\includegraphics[width=0.52\linewidth, height=0.22\textheight]{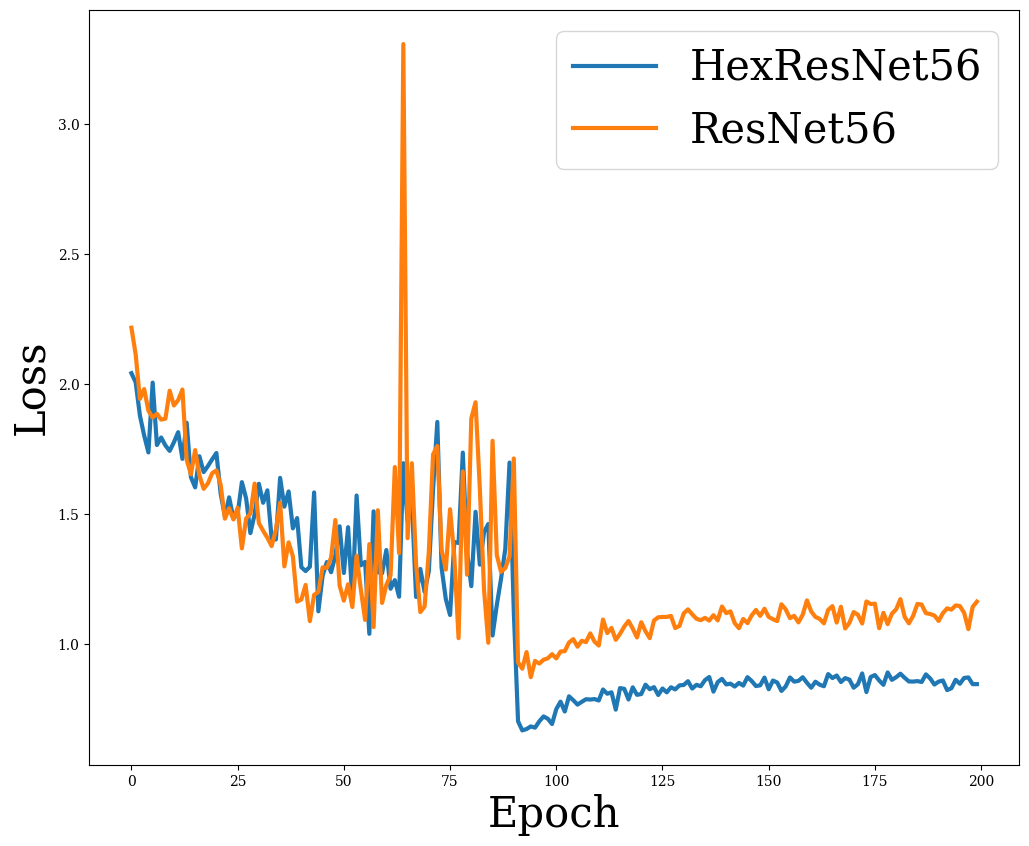}   \\
		(d)     \\
	\end{tabular}
	\caption{Validation Loss variation with respect to epochs for ImageNet dataset. ResNet vs HexResNet: (a) 20 layers, (b) 32 layers, (c) 44 layers, and (d) 56 layers.}
	\label{fig:12_a}
\end{figure}

\begin{table}[H]
	\centering
	\caption{Error rates and accuracy percentage on ImageNet dataset.}
	\begin{adjustbox}{width=0.8\textwidth}
		\label{tab:3}
		\begin{tabular}{|c|c|c|c|c|c|c|}
			\hline
			\textbf{Model} & \textbf{Parameters} & \textbf{\begin{tabular}[c]{@{}c@{}}Validation \\ Acc\end{tabular}} & \textbf{\begin{tabular}[c]{@{}c@{}}Top1 \\ acc \%\end{tabular}} & \textbf{\begin{tabular}[c]{@{}c@{}}Top 1 \\ error\%\end{tabular}} & \textbf{\begin{tabular}[c]{@{}c@{}}Top 5 \\ acc \%\end{tabular}} & \textbf{\begin{tabular}[c]{@{}c@{}}Top 5 \\ error \%\end{tabular}} \\ \hline
			ResNet - 20 & 272, 474 & 82.  84\% & 9.  40\% & 90.  60\% & 52.  80\% & 47.  19\% \\ \hline
			\textbf{ResNet - 20(H)} & \textbf{287, 130} & \textbf{83.    61\%} & \textbf{10.    60\%} & \textbf{89.    40\%} & \textbf{52.    20\%} & \textbf{47.    80\%} \\ \hline
			ResNet - 32 & 466, 906 & 78.  64\% & 10.  20\% & 89.  80\% & 49.  60\% & 50.  54\% \\ \hline
			\textbf{ResNet - 32(H)} & \textbf{481, 114} & \textbf{84.    49\%} & \textbf{11.    40\%} & \textbf{88.    60\%} & \textbf{51.    60\%} & \textbf{48.    40\%} \\ \hline
			ResNet - 44 & 661, 338 & 82.  56\% & 10.  80\% & 89.  20\% & 49.  00\% & 51.  00\% \\ \hline
			\textbf{ResNet - 44(H)} & \textbf{675, 098} & \textbf{80.    60\%} & \textbf{12.    80\%} & \textbf{87.    20\%} & \textbf{50.    60\%} & \textbf{49.    40\%} \\ \hline
			ResNet - 56 & 855, 770 & 80.  60\% & 8.  60\% & 91.  40\% & 49.  20\% & 50.  80\% \\ \hline
			\textbf{ResNet - 56(H)} & \textbf{869, 082} & \textbf{83.    50}\% & \textbf{10.    60}\% & \textbf{89.    40\%} & \textbf{52.    20\%} & \textbf{47.    80\%} \\ \hline
		\end{tabular}
	\end{adjustbox}
\end{table}

We present the quantitative analysis correspond to ImageNet dataset in Table. \ref{tab:3}. The results in Table \ref{tab:3} show that proposed Hex-ResNet configurations has lesser validation error compared to their respective ResNet models. Also,  Hex-ResNet 34 is better than ResNet 20 layer with significantly lower training error. This shows that the vanishing gradient problem is more well addressed and acquired high accuracy percentage with the proposed Hex-ResNet architecture.  Secondly, compared to the classical ResNet counterparts, the Hex-ResNet 20, 32, 44, and 56 layer reduces the Top-1 error leading to reduced training error. This is one of the significant comparisons that depicts the effectiveness of hex residual learning on extremely deeper systems. Thirdly, Hex-ResNet bring down the top-1 error by 1.2\%  in lower ResNet models (20,32) and 2\% in higher ResNet models (44,56). This comparison clearly shows the benefit of hex residual learning, especially on deeper network systems. Lastly, Figs. \ref{fig:12_a} (a)-(d) shows the variation of validation loss with respect to epochs for various ResNet configurations. It can be seen that the Hex-ResNet architecture converges faster as compared to the classical ResNet. Hence, Hex-ResNet has faster and more accurate convergence at the initial stages itself. A slight exception could be found with the Hex-ResNet 56 which performed slightly worse than the classical ResNet in the case of validation error. The probable reason behind this outlier behavior is due to the fact that the HexResNet56 is overfitting the ImageNet data. Also, the stability of validation loss variation is much better than its corresponding ResNet counterparts. This indicates that our approach is less prone to noisy data than when compared to ResNet defined on pure square tessellations. Most importantly, it can be seen from both the Tables. \ref{tab:2} and \ref{tab:3} that the number of additional parameters due to hexagonal convolutions is minimal when compared to the number of parameters in their respective ResNet counterparts.

\section{CONCLUSION}
In this research work, we aimed at a new novel architecture for ResNet which replaces the existing square convolution by hexagonal convolution in skip connection for challenging computer vision tasks.  We have integrated the hexagonal convolution and the existing training procedures under distinct resource limitations. We have reported consistent improvement on baseline architecture.  From the experimental results, we could observe that the top-5 accuracy of Hex-ResNet is outperforming classical ResNet indicating its generalization capabilities.  It also shows that Hex-ResNet can learn lot more details even with a smaller dataset, providing higher accuracy, less error with classical ResNet in a similar dataset. For instance, on the deep Hex- ResNet variants, we present improvements in Top-1 error and Top -5 accuracy,  ranging from 2\% to 1.2\% and 0.48\%  to 1.35\%. These improvements are achieved without increasing the model complexity.      
Future works can be extended to the following two tasks:
1) to different computer vision applications with improved training procedures.
2) to different deep neural networks to be trained via Hex-convolutions.

%Bibliography
\bibliographystyle{unsrt}  
\bibliography{references}

\begin{thebibliography}{10}

\bibitem{heresnet}
Kaiming He, Xiangyu Zhang, Shaoqing Ren, and Jian Sun.
\newblock Deep residual learning for image recognition.
\newblock In {\em Proceedings of the IEEE conference on computer vision and
  pattern recognition}, pages 770--778, 2016.

\bibitem{imagenet_cnn}
Alex Krizhevsky, Ilya Sutskever, and Geoffrey~E Hinton.
\newblock Imagenet classification with deep convolutional neural networks.
\newblock {\em Advances in neural information processing systems},
  25:1097--1105, 2012.

\bibitem{zeiler_visual}
Matthew~D Zeiler and Rob Fergus.
\newblock Visualizing and understanding convolutional networks.
\newblock In {\em European conference on computer vision}, pages 818--833.
  Springer, 2014.

\bibitem{alom2018history}
Md~Zahangir Alom, Tarek~M Taha, Christopher Yakopcic, Stefan Westberg, Paheding
  Sidiki, Mst~Shamima Nasrin, Brian~C Van~Essen, Abdul A~S Awwal, and Vijayan~K
  Asari.
\newblock The history began from alexie: A comprehensive survey on deep
  learning approaches.
\newblock {\em arXiv preprint arXiv:1803.01164}, 2018.

\bibitem{bengiograd}
Yoshua Bengio, Patrice Simard, and Paolo Frasconi.
\newblock Learning longterm dependencies with gradient descent is difficult.
\newblock {\em IEEE transactions on neural networks}, 5(2):157--166, 1994.

\bibitem{xavier_grad}
Xavier Glorot and Yoshua Bengio.
\newblock Understanding the difficulty of training deep feedforward neural
  networks.
\newblock In {\em Proceedings of the thirteenth international conference on
  artificial intelligence and statistics}, pages 249--256. JMLR Workshop and
  Conference Proceedings, 2010.

\bibitem{harder_opt}
Mohammad~Sadegh Ebrahimi and Hossein~Karkeh Abadi.
\newblock Study of residual networks for image recognition.
\newblock In {\em Intelligent Computing}, pages 754--763. Springer, 2021.

\bibitem{wu2019wider}
Zifeng Wu, Chunhua Shen, and Anton Van Den~Hengel.
\newblock Wider or deeper: Revisiting the resnet model for visual recognition.
\newblock {\em Pattern Recognition}, 90:119--133, 2019.

\bibitem{farooq2020covid}
Muhammad Farooq and Abdul Hafeez.
\newblock Covid-resnet: A deep learning framework for screening of covid19 from
  radiographs.
\newblock {\em arXiv preprint arXiv:2003.14395}, 2020.

\bibitem{heximages_online}
\url{www.pexels.com}.

\bibitem{curcio1990human}
Christine~A Curcio, Kenneth~R Sloan, Robert~E Kalina, and Anita~E Hendrickson.
\newblock Human photoreceptor topography.
\newblock {\em Journal of comparative neurology}, 292(4):497--523, 1990.

\bibitem{middleton}
Lee Middleton and Jayanthi Sivaswamy.
\newblock {\em Hexagonal image processing: A practical approach}.
\newblock Springer Science \& Business Media, 2006.

\bibitem{sheridan}
Phil Sheridan, Tom Hintz, and David Alexander.
\newblock Pseudo-invariant image transformations on a hexagonal lattice.
\newblock {\em Image and Vision Computing}, 18(11):907--917, 2000.

\bibitem{Mersereau}
Russell~M Mersereau.
\newblock The processing of hexagonally sampled two-dimensional signals.
\newblock {\em Proceedings of the IEEE}, 67(6):930--949, 1979.

\bibitem{mostafa}
K~Mostafa, JY~Chiang, and I~Her.
\newblock Edge-detection method using binary morphology on hexagonal images.
\newblock {\em The Imaging Science Journal}, 63(3):168--173, 2015.

\bibitem{Neurohex}
Kenny Young, Gautham Vasan, and Ryan Hayward.
\newblock Neurohex: A deep q-learning hex agent.
\newblock In {\em Computer Games}, pages 3--18. Springer, 2016.

\bibitem{iact}
Idan Shilon, Manuel Kraus, Matthias B{\"u}chele, Kathrin Egberts, Tobias
  Fischer, Tim~Lukas Holch, Thomas Lohse, Ullrich Schwanke, Constantin Steppa,
  and Stefan Funk.
\newblock Application of deep learning methods to analysis of imaging
  atmospheric cherenkov telescopes data.
\newblock {\em Astroparticle Physics}, 105:44--53, 2019.

\bibitem{icecube}
Mirco Huennefeld.
\newblock Deep learning in physics exemplified by the reconstruction of
  muon-neutrino events in icecube.
\newblock {\em Verhandlungen der Deutschen Physikalischen Gesellschaft}, 2017.

\bibitem{hoogeboom2018hexaconv}
Emiel Hoogeboom, Jorn~WT Peters, Taco~S Cohen, and Max Welling.
\newblock Hexaconv.
\newblock {\em arXiv preprint arXiv:1803.02108}, 2018.

\bibitem{hexagdly}
Constantin Steppa and Tim~L Holch.
\newblock Hexagdly—processing hexagonally sampled data with cnns in pytorch.
\newblock {\em SoftwareX}, 9:193--198, 2019.

\bibitem{resnetapp1}
Heechul Jung, Min-Kook Choi, Jihun Jung, Jin-Hee Lee, Soon Kwon, and Woo
  Young~Jung.
\newblock Resnet-based vehicle classification and localization in traffic
  surveillance systems.
\newblock In {\em Proceedings of the IEEE conference on computer vision and
  pattern recognition workshops}, pages 61--67, 2017.

\bibitem{resnetapp2}
Tongyan Gong and Huiqian Niu.
\newblock An implementation of resnet on the classification of rgb-d images.
\newblock In {\em International Symposium on Benchmarking, Measuring and
  Optimization}, pages 149--155. Springer, 2019.

\bibitem{baoqi}
Baoqi Li and Yuyao He.
\newblock An improved resnet based on the adjustable shortcut connections.
\newblock {\em IEEE Access}, 6:18967--18974, 2018.

\bibitem{resnet_timm}
Ross Wightman, Hugo Touvron, and Herv{\'e} J{\'e}gou.
\newblock Resnet strikes back: An improved training procedure in timm.
\newblock {\em arXiv preprint arXiv:2110.00476}, 2021.

\bibitem{identity_mappings}
Kaiming He, Xiangyu Zhang, Shaoqing Ren, and Jian Sun.
\newblock Identity mappings in deep residual networks.
\newblock In {\em European conference on computer vision}, pages 630--645.
  Springer, 2016.

\bibitem{hexadv1}
Daniel~P Petersen and David Middleton.
\newblock Sampling and reconstruction of wave-number-limited functions in
  n-dimensional euclidean spaces.
\newblock {\em Information and control}, 5(4):279--323, 1962.

\bibitem{hexadv2}
Russell~M Mersereau.
\newblock The processing of hexagonally sampled two-dimensional signals.
\newblock {\em Proceedings of the IEEE}, 67(6):930--949, 1979.

\bibitem{hexadv3}
Marcel~JE Golay.
\newblock Hexagonal parallel pattern transformations.
\newblock {\em IEEE Transactions on computers}, 100(8):733--740, 1969.

\bibitem{hexadv4}
Mohammad~Bagher Nourian and MR~Aahmadzadeh.
\newblock Image de-noising with virtual hexagonal image structure.
\newblock In {\em 2013 First Iranian Conference on Pattern Recognition and
  Image Analysis (PRIA)}, pages 1--5. IEEE, 2013.

\bibitem{hexadv5}
Sonya Coleman, Bryan Scotney, and Bryan Gardiner.
\newblock Tri-directional gradient operators for hexagonal image processing.
\newblock {\em Journal of Visual Communication and Image Representation},
  38:614--626, 2016.

\bibitem{hexadv6}
Illa Singh and Ashish Oberoi.
\newblock Comparison between square pixel structure and hexagonal pixel
  structure in digital image processing.
\newblock {\em Int. J. Comput. Sci. Trends Technol}, 3:176--181, 2015.

\bibitem{van2004hex}
Dimitri Van De~Ville, Thierry Blu, Michael Unser, Wilfried Philips, Ignace
  Lemahieu, and Rik Van~de Walle.
\newblock Hex-splines: A novel spline family for hexagonal lattices.
\newblock {\em IEEE Transactions on Image Processing}, 13(6):758--772, 2004.

\bibitem{middleton2001edge}
Lee Middleton and Jayanthi Sivaswamy.
\newblock Edge detection in a hexagonal-image processing framework.
\newblock {\em Image and Vision computing}, 19(14):1071--1081, 2001.

\bibitem{abbas2015pet}
Syed~Tabish Abbas and Jayanthi Sivaswamy.
\newblock Pet image reconstruction and denoising on hexagonal lattices.
\newblock In {\em 2015 IEEE International Conference on Image Processing
  (ICIP)}, pages 3481--3485. IEEE, 2015.

\bibitem{contreras2014hexagonal}
Sonia~H Contreras-Ortiz and Martin~D Fox.
\newblock Hexagonal filters for ultrasound images.
\newblock {\em Journal of Electronic Imaging}, 23(4):043022, 2014.

\bibitem{ortiz2011hexagonal}
Sonia H~Contreras Ortiz, Tsuicheng Chiu, and Martin~D Fox.
\newblock Hexagonal adaptive filtering on compound ultrasound images.
\newblock In {\em 2011 Annual International Conference of the IEEE Engineering
  in Medicine and Biology Society}, pages 4856--4859. IEEE, 2011.

\bibitem{hexdev1}
Do-Hyeong Kim, Munkh-Uchral Erdenebat, Ki-Chul Kwon, Ji-Seong Jeong, Jae-Won
  Lee, Kyung-Ah Kim, Nam Kim, and Kwan-Hee Yoo.
\newblock Real-time 3d display system based on computer-generated integral
  imaging technique using enhanced ispp for hexagonal lens array.
\newblock {\em Applied Optics}, 52(34):8411--8418, 2013.

\bibitem{hexdev2}
Amir Hassanfiroozi, Yi-Pai Huang, Bahram Javidi, and Han-Ping~D Shieh.
\newblock Hexagonal liquid crystal lens array for 3d endoscopy.
\newblock {\em Optics express}, 23(2):971--981, 2015.

\bibitem{hexdev3}
Ayatollah Karimzadeh.
\newblock Analysis of the depth of field in hexagonal array integral imaging
  systems based on modulation transfer function and strehl ratio.
\newblock {\em Applied Optics}, 55(11):3045--3050, 2016.

\bibitem{zhao2020hexcnn}
Yunxiang Zhao, Qiuhong Ke, Flip Korn, Jianzhong Qi, and Rui Zhang.
\newblock Hexcnn: A framework for native hexagonal convolutional neural
  networks.
\newblock In {\em 2020 IEEE International Conference on Data Mining (ICDM)},
  pages 1424--1429. IEEE, 2020.

\bibitem{hexdnn1}
Tobias Schlosser, Michael Friedrich, and Danny Kowerko.
\newblock Hexagonal image processing in the context of machine learning:
  Conception of a biologically inspired hexagonal deep learning framework.
\newblock In {\em 2019 18th IEEE International Conference on Machine Learning
  and Applications (ICMLA)}, pages 1866--1873. IEEE, 2019.

\bibitem{hexdnn2}
Junren Luo, Wanpeng Zhang, Jiongming Su, and Fengtao Xiang.
\newblock Hexagonal convolutional neural networks for hexagonal grids.
\newblock {\em IEEE Access}, 7:142738--142749, 2019.

\bibitem{chen2020deep}
Dongwei Chen, Fei Hu, Guokui Nian, and Tiantian Yang.
\newblock Deep residual learning for nonlinear regression.
\newblock {\em Entropy}, 22(2):193, 2020.

\bibitem{improved_resnet}
Baoqi Li and Yuyao He.
\newblock An improved resnet based on the adjustable shortcut connections.
\newblock {\em IEEE Access}, 6:18967--18974, 2018.

\bibitem{cifar}
Alex Krizhevsky and Geoffrey Hinton.
\newblock Learning multiple layers of features from tiny images.
\newblock {\em IEEE Transactions on Power Electronics}, 28(11):5049--5062,
  2009.

\end{thebibliography}

\end{document}